\documentclass[10pt,twocolumn,letterpaper]{article}

\usepackage{cvpr}
\usepackage{times}
\usepackage{epsfig}
\usepackage{graphicx}
\usepackage{amsmath}
\usepackage{amssymb}
\usepackage{authblk}
\usepackage{multirow}
\usepackage{booktabs} % Allows the use of \toprule, \midrule and
\usepackage{algorithm}
\usepackage{algorithmic}

\usepackage{adjustbox}
\usepackage{booktabs}
\usepackage{multirow}
\usepackage{subcaption}
\usepackage{tabularx}
\usepackage{float}
\usepackage{pifont}
\newcommand{\xmark}{\ding{55}}%

\usepackage[pagebackref=true,breaklinks=true,letterpaper=true,colorlinks,bookmarks=false]{hyperref}

\cvprfinalcopy % *** Uncomment this line for the final submission

\def\cvprPaperID{2956} % *** Enter the CVPR Paper ID here

\ifcvprfinal\fi
\begin{document}

%%%%%%%%% TITLE
\title{Single View Stereo Matching}

\author{
Yue Luo$^1$ $^*$ \hspace{0.03in} Jimmy Ren$^1$ \thanks{Indicates equal contribution.} \hspace{0.03in} Mude Lin$^1$ \hspace{0.03in} Jiahao Pang$^1$ \hspace{0.03in} Wenxiu Sun$^1$ \hspace{0.03in} Hongsheng Li$^2$ \hspace{0.03in} Liang Lin$^{1,3}$\\
$^1$SenseTime Research\\
$^2$The Chinese University of Hong Kong, Hong Kong SAR, China\\
$^3$Sun Yat-sen University, China\\
$^1$\{luoyue,rensijie,linmude,pangjiahao,sunwenxiu,linliang\}@sensetime.com\\
$^2$hsli@ee.cuhk.edu.hk\\
}

\maketitle
\thispagestyle{empty}

%%%%%%%%% ABSTRACT
\begin{abstract}
   Previous monocular depth estimation methods take a single view and directly regress the expected results. Though recent advances are made by applying geometrically inspired loss functions during training, the inference procedure does not explicitly impose any geometrical constraint. Therefore these models purely rely on the quality of data and the effectiveness of learning to generalize. This either leads to suboptimal results or the demand of huge amount of expensive ground truth labelled data to generate reasonable results. In this paper, we show for the first time that the monocular depth estimation problem can be reformulated as two sub-problems, a view synthesis procedure followed by stereo matching, with two intriguing properties, namely i) geometrical constraints can be explicitly imposed during inference; ii) demand on labelled depth data can be greatly alleviated. We show that the whole pipeline can still be trained in an end-to-end fashion and this new formulation plays a critical role in advancing the performance. The resulting model outperforms all the previous monocular depth estimation methods as well as the stereo block matching method in the challenging KITTI dataset by only using a small number of real training data. The model also generalizes well to other monocular depth estimation benchmarks. We also discuss the implications and the advantages of solving monocular depth estimation using stereo methods.\footnote{Code is publicly available at \url{https://github.com/lawy623/SVS}.}
\end{abstract}

%%%%%%%%% BODY TEXT
\vspace{-5pt}
\section{Introduction}
Depth estimation is one of the fundamental problems in computer vision. It finds important applications in a large number of areas such as robotics, augmented reality, 3D reconstruction and self-driving car, etc. This problem is heavily studied in the literature and is mainly tackled with two types of technical methodologies namely active stereo vision such as structured light \cite{ScharsteinS03light}, time-of-flight \cite{ZhuWYDP11TOF}, and passive stereo vision including stereo matching\cite{lad2015stereo,luo2016stereo}, structure from motion \cite{SturmT96sfm}, photometric stereo \cite{EstebanVC08} and depth cue fusion \cite{Saxena09make3D}, etc. Among passive stereo vision methods, stereo matching is arguably the most widely applicable technique because it is accurate and it poses little assumption to the sensors and the imaging procedure. Recent advances in this field show that the quality of stereo matching can be significantly improved by deep models trained with synthetic data and finetuned with limited amount real data \cite{mayer2016disp,pang2017cascade}. 

\begin{figure}[t!]
  \includegraphics[width=8.2cm,height=3.4cm]{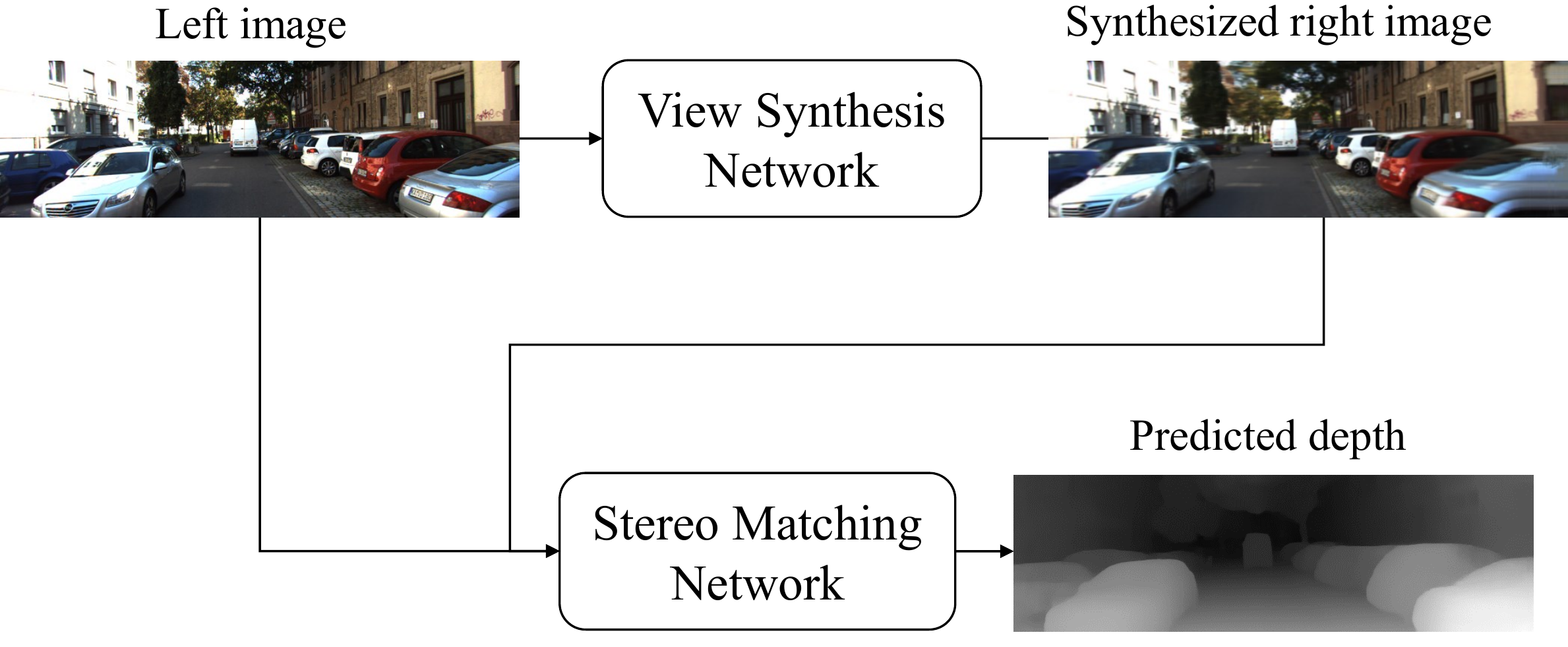}
  \vspace{-5pt}
  \caption{Pipeline of our approach on monocular depth estimation. We decompose the task into two parts: view synthesis and stereo matching. Both networks enforce the geometric reasoning capacity. With this new formulation, our approach is able to achieve state-of-the-art performance.}
  \vspace{-13pt}
  \label{figure1}
\end{figure}

On the other hand, the applicability of monocular depth estimation is greatly limited by its accuracy though the single camera setting is much more preferred in practice in order to avoid calibration errors and synchronization problems occur to the stereo camera setting. Estimating depth from a single view is difficult because it is an ill-posed and geometrically ambiguous problem. Advancement of monocular depth estimation has recently been made by deep learning methods \cite{eigen2014depth,laina2016deeper,li2015depth,liu2015deep}
. However, comparing to the mentioned passive stereo vision methods which are grounded by geometric correctness, the formulation in the current state-of-the-art monocular method is problematic. The reasons are twofold. First, current deep learning approaches to this problem almost completely rely on the high-level semantic information and directly relate it to the absolute depth value. Because the operations in the network are general and do not have any prior knowledge on the function it needs to approximate, learning such semantic information is difficult even some special constraints are imposed in the loss function. Second, even the effective learning can be achieved, the relationship between scene understanding and depth needs to be established by a huge number of real data with ground truth depth. Such data is not only very expensive to obtain at scale, collecting high-quality dense labels is very difficult and time consuming if not entirely impossible. This significantly limits the potential of the current formulation.

In this paper, we take a novel perspective and show for the first time that monocular depth estimation problem can be formulated as a stereo matching problem in which the right view is automatically generated by a high-quality view synthesis network. The whole pipeline is shown in figure \ref{figure1}. The key insights here are that i) both view synthesis and stereo matching respect the underlying geometric principles; ii) both of them can be trained without using the expensive real depth data and thus generalize well; iii) the whole pipeline can be collectively trained in an end-to-end fashion that optimize the geometrically correct objectives. Our method shares a similar idea as revealed in the Spatial Transformation Network \cite{Jaderberg2015STN}. Although deep models can learn necessary transformations by themselves, it might be more beneficial for us to explicitly model such transformations. We discover that the resulting model is able to outperform all the previous methods in the challenging KITTI dataset \cite{Geiger2012CVPR} by only using a small number of real training data. The model also generalizes well to other monocular depth estimation datasets.

Our contributions can be summarized as follows.
\begin{itemize}
\item First, we discover that the monocular depth estimation problem can be effectively decoupled into two sub-problems with geometrical soundness. It forms a new foundation in advancing the performance in this field.
\item Second, we show that the whole pipeline can be trained end-to-end and it outperforms all the previous monocular methods by a large margin using a fraction of training data. Notably, this is the first monocular method to outperform the stereo blocking matching algorithm in terms of the overall accuracy.
\end{itemize}

\section{Related Works}
There exists a large body of literature on depth estimation from images, either using single view~\cite{saxena20083}, stereo views~\cite{scharstein2002taxonomy}, several overlapped images from different viewpoints~\cite{furukawa2015multi}, or temporal sequence~\cite{ranftl2016dense}.
For monocular depth estimation, Saxena \etal~\cite{saxena20083} propose one of the first supervised learning-based approaches to single image depth map prediction. They model depth prediction in a Markov random field and use multi-scale texture features that have been hand-crafted. Recently, deep learning has proven its ability in many computer vision tasks, including the single image depth estimation. Eigen \etal~\cite{eigen2014depth} propose the first CNN framework that predicts the depth in a coarse-to-fine manner. Laina \etal~\cite{laina2016deeper} employ a deeper ResNet~\cite{He2015res} structure with an efficient up-sampling design and achieve a boosted performance. Liu \etal~\cite{liu2015deep} also propose a deep structured learning approach that allows for training CNN features of unary and pairwise potentials in an end-to-end way. Chen \etal~\cite{chen2016wild} provide a novel insight by incorporating pair-wise depth relation into CNN training. Compared with depth, these rankings on pixel level are much more easy to obtain. Further lines of research in supervised training of depth map prediction use the idea of depth transfer from example images~\cite{karsch2012depth, konrad20122d, liu2014discrete}, or combining semantic segmentation~\cite{eigen2015predicting,ladicky2014pulling,li2010towards,liu2010single,
wang2015towards}. However, large amount of high-quality labels are in need to establish the transformation from image space to depth space. Such data are not easy to collect at scale in real life.  

Recently, a small number of deep network based methods attempt to estimate depth in an unsupervised way. Garg \etal~\cite{garg2016unsupervised} first introduce the unsupervised method by only supervising on the image alignment loss. However, their loss is not fully differentiable so that they apply first Taylor expansion to linearize their loss for back-propagation. Godard \etal~\cite{godard2016unsupervised} also propose an unsupervised deep learning framework, and they employ a novel loss function to enforce consistency between the predicted depth maps from each camera view. Kuznietsov \etal~\cite{kuznietsov2017semi} adopt a semi-supervised deep method to predict depths from single images. Sparse depth from LiDAR sensors is used for supervised learning, while a direct image alignment loss is integrated to produce photoconsistent dense depth maps in a stereo setup. Zhou \etal~\cite{zhou2017unsupervised} jointly estimate depth and camera pose in an unsupervised manner.

Despite that those unsupervised methods reduce the demand of expensive depth ground truth, their mechanisms are still inherently problematic since they are attempting to regress a depth/disparity directly from a single image. The network architecture itself does not assume any geometric constraints and it acts like a black box. In our work, we propose a novel strategy to decompose this task into two separate procedures, namely synthesizing a corresponding right view followed by a stereo matching procedure. Such idea is similar to the Spatial Transformation Network \cite{Jaderberg2015STN}, which learns a transformation within the network before conducting visual tasks like recognition.

To synthesize a novel view, DeepStereo ~\cite{flynn2015deepStereo} first proposes to render an unseen view by taking pixels from other views, and \cite{zhou2016view} predicts the appearance flow to reconstruct the target view. The Deep3D network of Xie \etal~\cite{xie2016deep3d} addresses the problem of generating the corresponding right view from an input left image. Their method produces a distribution over all the possible disparities for each pixel, which is used to generate the right image.

\begin{figure*}[t!]
  \includegraphics[width=17.4cm,height=6.5cm]{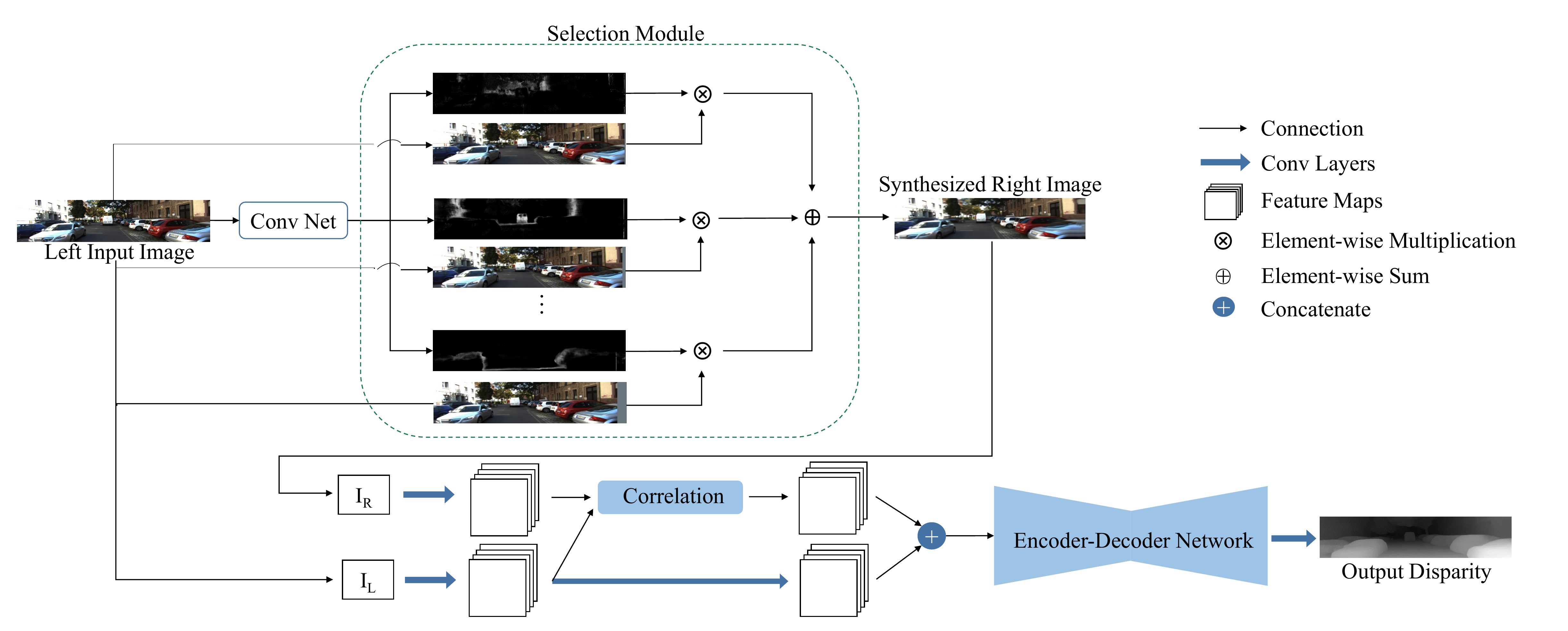}
  \vspace{-20pt}
  \caption{Details of our single view stereo matching network. Upper part is the view synthesis network. The input image is first processed by a CNN. It results in probabilistic disparity maps that help to reconstruct a synthetic right view by selectively taking pixels from nearby locations on the original left image. A stereo matching network, which is shown on the lower part of the figure, then takes both the original left image and synthetic right image to calculate an accurate disparity, which can be transformed into a corresponding depth map given the camera settings.}
  \label{figure2}
  \vspace{-15pt}
\end{figure*}

Conducting stereo matching on the original left input and the synthetic right view is now a 1D matching problem. The vast majority of works on stereo matching focus on learning a matching function that searches the corresponding pixels on two images \cite{lad2015stereo,luo2016stereo}. Mayer \etal~\cite{mayer2016disp} introduce their fully convolutional DispNet to directly regress the disparity from the stereo pair. Later, Pang \etal~\cite{pang2017cascade} adopt a multi-scale residual network developed from DispNet and obtain refined results. These methods still rely on large amount labelled disparity as ground truth. Instead of using data from the real world, training on synthetic data \cite{mayer2016disp} becomes a more feasible solution to these approaches.

\section{Analysis and our approach}
In this section, we demonstrate how we decompose the task of monocular depth estimation into two separate tasks. And we illustrate our model design for view synthesis and stereo matching separately. 

\subsection{Analysis of the whole pipeline}
In our pipeline, we decompose the task of monocular depth estimation into two tasks, namely view synthesis and stereo matching. The whole pipeline is shown in figure \ref{figure2}. By tackling this problem using two separate steps, we find that both procedures obey primary geometric principles and they can be trained without expensive data supply. After that, these networks can be collectively trained in an end-to-end manner. We further hypothesize that, when the whole pipeline is trained end-to-end, both components will not degrade their capacity of constraining geometric correctness, and the performance of the whole pipeline will be promoted thanks to joint training. Therefore, we are desired to choose both methods that can explicitly model the geometric transformation in the network design.

The first stage is view synthesis. For a stereo pair, binocular views are rendered by well synchronized and calibrated cameras, resulting in the strong correspondence between pixels in the horizontal direction. Unlike previous warp-based methods that generally require an accurate estimation of the underlying geometry, Deep3D ~\cite{xie2016deep3d} proposes a new probabilistic scheme to transfer pixels from the original image. By this mean, it directly formulates the transformation from left image to right image using a differentiable selection layer. We adopt its design and develop our view synthesis network based on it. Other reconstruction plans \cite{garg2016unsupervised,godard2016unsupervised,kuznietsov2017semi} are also viable alternatives, but the choice of the specific view synthesis method is independent of the main insight of the paper.

After generating a high-quality novel view, our stereo matching network transforms the high-level scene understanding problem into a 1D matching problem, which results in less computational complexity. In order to better utilize the geometric relation between two views, we take the idea of 1D correlation employed in DispNetC\cite{mayer2016disp}. We further adopt the DispFullNet structure mentioned in \cite{pang2017cascade} to achieve full resolution prediction.

\subsection{View synthesis network}
Our view synthesis network is shown in the upper part of figure \ref{figure2}. We develop this network based on Deep3D ~\cite{xie2016deep3d} model. Here we briefly introduce the structure of it. At the very beginning, an input left image $I_{l}$ is processed by a baseline network. We then upsample the features from different intermediate levels to the same resolution, in order to incorporate low-level features into final use. Those features are then summed up to further produce a probabilistic disparity map. After completing a selection operation, pixels on original $I_{l}$ can be selectively mixed up to form a new pixel on the right image.

The operation of selection is the core component in this network. This module is also illustrated in figure \ref{figure2}. Denote $I_{l}$ as the input left image, previous Depth Image-Based Rendering (DIBR) techniques choose to directly warp the left image based on estimated disparity into a corresponding right image. Suppose $D$ is the predicted disparity aligned with the left image, the procedure can be formulated as

\vspace{-8pt}
\begin{equation}
\label{DIBR}
\begin{aligned}
\widetilde{I}_{r}(i,j - D_{i,j})&=I_{l}(i,j),&\quad (i,j)\in{\Omega_{l}},&\\
\end{aligned}
\end{equation}
\vspace{-13pt}

where $\Omega_{l}$ is the image space of $I_{l}$ and $i$, $j$ refer to the row and column on $I_{l}$ respectively. Though this function captures the geometric correspondence between images in a stereo setup, it requires an accurate disparity map to reconstruct the right view. At the same time, the function is not fully differentiable with respect to $D$ which limits the opportunity of training by a deep neural network. The selection module, instead, formulates the reconstruction as a process of probabilistic summation. Denote $D\in\mathbb{R}^{W\times H\times C}$ as the probabilistic disparity result, where $W$ and $H$ are the width and height of left image and 
$C$ indicates the number of possible disparity shifts, the reconstruction can then be formulated as

\vspace{-11pt}
\begin{equation}
\label{selection}
\begin{aligned}
\widetilde{I}_{r}&=\sum_d I_{l}^{(d)}D^d.
\end{aligned}
\end{equation}
\vspace{-9pt}

Here, $I_{l}^{(d)}(i,j) = I_{l}(i,j+d)$ is the shifted left image whose stride is predetermined by possible disparity values $d$. This operation sums up the stacked shifted input by learned weights and ensures the differentiability of the whole system.

To supervise the reconstruction quality, we do not propose any special loss function. We find that a simple L1 loss supervising on the reconstructed appearance is sufficient for the task of view synthesis:

\vspace{-8pt}
\begin{equation}
\label{L1Loss}
\begin{aligned}
L_{view} = \frac{1}{N}\sum_{i,j}\left| \widetilde{I}_{r}(i,j) - I_{r}(i,j)\right|
\end{aligned}
\end{equation}

\subsection{Stereo matching network}
There exists a large body of literature tackling the problem of stereo matching. Recent advancements are achieved by deep learning models. Not only because deep networks help to effectively find out similar pixel pairs, research also show that these networks can be trained on a large amount of synthetic data and they can still generalize well on real images \cite{mayer2016disp}. In our pipeline, we select the state-of-the-art DispNetC \cite{mayer2016disp} structure as the desired network for the stereo matching task. We further follow the modifications made in \cite{pang2017cascade} to adopt a DipFulNet structure for full-resolution output. The structure of this method can be seen in the lower part of figure \ref{figure2}. We briefly illustrate the method here, and the detailed settings can be found in their papers.

After processed by several convolutional operations, 1D correlation will be calculated based on resulted features. This correlation layer is found very useful in the stereo matching problem since it explicitly encodes the geometric relationship into the model design, and the horizontal correlation is indeed an effective cue for finding the most similar pairs. The features will be further concatenated with higher-level features from the left image $I_{l}$. An encoder-decoder network further processes the concatenated features and produces disparity at different scales. These intermediate and final results will be supervised by ground truth disparity using L1 loss. 

\subsection{End-to-end training of the whole pipeline}

These two networks can be combined for joint training once being trained to obtain the ability of geometric reasoning for the task of view synthesis and stereo matching separately. End-to-end training of the whole pipeline can thus be performed to enforce the collaboration of these two sub-networks.

\section{Experiments}
In this section, we present our experiments and results. Our method achieves state-of-the-art monocular depth estimation result on the widely used KITTI dataset \cite{Geiger2012CVPR}. We discover and show the key insights of this method and prove the correctness of our methodology. We also make the first attempt to run our single view approach on the challenging KITTI Stereo 2015 benchmark \cite{Menze2015CVPR}.

\subsection{Dataset and Evaluation Metrics}
We evaluate our approach on the publicly available KITTI benchmark \cite{Geiger2012CVPR}. In order to fairly compare with other methods on monocular depth estimation, we use the raw sequences of KITTI and employ the split scheme proposed by Eigen \etal~\cite{eigen2014depth}. This split results in a test set with 697 images. Remaining data is used for training and validation. Overall we have 22600 stereo pairs for training our view synthesis network. Except for stereo image pairs, the dataset also contains sparse 3D laser measurements taken from a Velodyne laser sensor. They can be projected onto image space and served as the depth labels. Parameters of the stereo setup and the camera intrinsics are also provided, therefore we can transfer depth into disparity as ground truth during end-to-end training and recover the depth from disparity during inference.

Evaluation metrics are as follows and they indicate the error and performance on predicted monocular depth.

ARD = $\frac{1}{N}\sum_{i=1}^N{\left|Dep_i - Dep_i^{g.t.}\right|}/{Dep_i^{g.t.}}$

SRD = $\frac{1}{N}\sum_{i=1}^N{\lVert|Dep_i - Dep_i^{g.t.}\rVert^2}/{Dep_i^{g.t.}}$

RMSE = $\sqrt{\frac{1}{N}\sum_{i=1}^N\lVert Dep_i - Dep_i^{g.t.} \rVert^2}$

RMSE(log) = $\sqrt{\frac{1}{N}\sum_{i=1}^N\lVert log(Dep_i) - log(Dep_i)^{g.t.} \rVert^2}$

Accuracy = \% $Dep_i$ : $max(\frac{Dep_i}{Dep_i^{g.t.}},\frac{Dep_i^{g.t.}}{Dep_i})=\delta < thr$

Here N is the number of pixels that are not empty on the depth ground truth.

To compare with other works in a consistent manner, we only evaluate on a cropped region proposed by Eigen \etal~\cite{eigen2014depth}. Also, previous methods restrict the depth distance in different ranges for evaluation, we provide our result using both the cap of 0-80m (following Eigen \etal~\cite{eigen2014depth}) and 1-50m (following Garg \etal~\cite{garg2016unsupervised}). This requires to discard the pixels on which the depth is outside the proposed range. 

\subsection{Implementation Details}
\label{secion:impl_details}
The training of the model is divided into two stages. First we train the two networks used for different purposes separately. In the second stage, we combine the two parts and further finetune the whole pipeline in an end-to-end fashion. The training is conducted using caffe framework \cite{jia14caffe}.

In the first stage, networks are trained separately. For the training of view synthesis network, 22600 stereo pairs from KITTI are taken into use. We select VGG16 as the baseline network and initialize the weights of it using the model pre-trained from ImageNet \cite{Simonyan14c}. All other weights are initialized following the same scheme in \cite{xie2016deep3d}. Compared with original deep3D model \cite{xie2016deep3d}, we make some modifications to make it suitable for view synthesis task on KITTI dataset. First, the size of input is larger and is selected to be $640\times192$. It retains the aspect ratio of original KITTI images. Second, one more convolution layer is employed before deconvolution at each branch. Third, since the disparity ranges differently in KITTI and 3D movie dataset, we change the possible disparity range. A 65-channel probabilistic map representing possible disparity from 0 to 64 now becomes the final features. Last, to accommodate larger inputs and the deeper network structure, we decrease the batch size as 2, and we remove the origin BatchNorm layers in the deep3D model. The model is trained for 200K iterations with initial learning rate equals to 0.0002. For the training of DispFullNet used for stereo matching, we follow the training scheme specified in \cite{pang2017cascade}. The model is trained mainly on the synthetic FlyingThings3D dataset \cite{mayer2016disp} and optional finetuned on the KITTI stereo training set \cite{Menze2015CVPR}. This KITTI stereo training set contains 200 stereo pairs with relatively high-quality disparity labels, and it has not overlap with the test data from KITTI Eigen test set. The detailed settings can be found in Pang's paper \etal~\cite{pang2017cascade}.

In the second stage, two networks with pre-trained weights are now trained end-to-end. A small number of data from the KITTI Eigen training set with ground truth disparity labels will be taken to finetune the whole pipeline. Since the input to the stereo matching network has a larger dimension, upsample is performed inside the network to enlarge the synthetic right view resulted from the first stage.

Data augmentation is optionally done in both stages. The input will be randomly resized to a dimension slightly greater than the desired input size. And then it will be cropped into the desired size and fed into the network. The color intensity will also multiply a factor between 0.8 to 1.2.

\begin{table*}[tp]
\setlength{\tabcolsep}{10pt}
	\footnotesize
	\centering
	\begin{adjustbox}{max width=1.0\textwidth}
	\begin{tabular}{@{}l|c|cccc|ccc@{}}
	\toprule
\multicolumn{1}{l|}{Approach} & \multicolumn{1}{c|}{cap} & ARD   & SRD   & RMSE   & RMSE(log)   & $\delta < 1.25$ & $\delta < 1.25^2$ & $\delta < 1.25^3$ \\ \cline{3-9} 
                             &                         & \multicolumn{4}{c|}{lower is better} & \multicolumn{3}{c}{higher is better} \\ \midrule
Stereo\_gt\_right                    & $0-80$ m & 0.062 & 0.424 & 3.677 & 0.164     & 0.939         & 0.968         & 0.981         \\ \midrule
Eigen \etal~\cite{eigen2014depth}                   & $0-80$ m & 0.215 & 1.515 & 7.156 & 0.270     & 0.692         & 0.899         & 0.967         \\
Liu \etal~\cite{liu2014discrete}                     & $0-80$ m & 0.217 & 1.841 & 6.986 & 0.289     & 0.647         & 0.882         & 0.961         \\
Zhou \etal~\cite{zhou2017unsupervised}                    & $0-80$ m & 0.183 & 1.595 & 6.709 & 0.270     & 0.734         & 0.902         & 0.959         \\ 
Godard \etal~\cite{godard2016unsupervised}                  & $0-80$ m & 0.114 & 0.898 & 4.935 & 0.206     & 0.861         & 0.949         & 0.976         \\
Kuznietsov \etal~\cite{kuznietsov2017semi}              & $0-80$ m & 0.113 & 0.741 & 4.621 & 0.189     & 0.862         & 0.960         & \textbf{0.986}         \\
Ours, w/o end-to-end finetuned & $0-80$ m& 0.102 & 0.700 & 4.681 & 0.200     & 0.872         & 0.954         & 0.978         \\
Ours                          & $0-80$ m& \textbf{0.094} & \textbf{0.626} & \textbf{4.252} & \textbf{0.177}     & \textbf{0.891}         & \textbf{0.965}         & 0.984         \\ \bottomrule
Stereo\_gt\_right                    & $1-50$ m & 0.058 & 0.316 & 2.675 & 0.152     & 0.947         & 0.971         & 0.983         \\ \midrule
Zhou \etal~\cite{zhou2017unsupervised}                              & $1-50$ m & 0.190 & 1.436 & 4.975 & 0.258     & 0.735         & 0.915         & 0.968         \\
Garg \etal~\cite{garg2016unsupervised}                    & $1-50$ m & 0.169 & 1.080 & 5.104 & 0.273     & 0.740         & 0.904         & 0.962         \\
Godard \etal~\cite{godard2016unsupervised}                  & $1-50$ m & 0.108 & 0.657 & 3.729 & 0.194     & 0.873         & 0.954         & 0.979        \\  
Kuznietsov \etal~\cite{kuznietsov2017semi}              & $1-50$ m & 0.108 & 0.595 & 3.518 &   0.179   & 0.875         & 0.964         & \textbf{0.988}         \\
Ours, w/o end-to-end finetuned & $1-50$ m & 0.097 & 0.539 & 3.503 & 0.187     & 0.885         & 0.960         & 0.981         \\
Ours                          & $1-50$ m & \textbf{0.090} & \textbf{0.499} & \textbf{3.266} & \textbf{0.167}     & \textbf{0.902}         & \textbf{0.968}         & 0.986         \\

	\bottomrule
	\end{tabular}
	\end{adjustbox}
	\vspace{-6pt}
	\caption{Quantitative results of our method and approaches reported in the literature on the test set of the KITTI Raw dataset used by Eigen \etal~\cite{eigen2014depth} for different caps on ground-truth and/or predicted depth. Best results are shown in bold. Our proposed method achieves improvement over all compared state-of-the-art approaches.}
	\label{table:kitti_res}
\vspace{-8pt}
\end{table*}

\subsection{Depth Estimation by Stereo Matching method}
First, the evaluation of depth estimation of the stereo matching network given perfect right images
is presented. The result is shown in the Table \ref{table:kitti_res}, denoted as ``Stereo\_gt\_right". The stereo matching network clearly outperforms state-of-the-art methods for single image depth estimation, even the stereo matching network is mainly trained on rendered dataset~\cite{mayer2016disp}. 

The intuition here is that predicting depth from stereo images has a much higher accuracy than predicting depth by any of the previous monocular depth methods. This means we are able to achieve much higher performance if we can provide a sophisticated view synthesis module.

\begin{table*}[tp]
	\centering
	\begin{adjustbox}{max width=1.0\textwidth}
	\begin{tabular}{@{}l|c|c|c|cccc|ccc@{}}
	\toprule
\multicolumn{1}{l|}{Approach} & \multicolumn{1}{c|}{FT VSN} & \multicolumn{1}{c|}{FT SMN}& \multicolumn{1}{c|}{cap} & ARD   & SRD   & RMSE   & RMSE(log)   & $\delta < 1.25$ & $\delta < 1.25^2$ & $\delta < 1.25^3$ \\ \cline{5-11} 
                             &               & &           & \multicolumn{4}{c|}{lower is better} & \multicolumn{3}{c}{higher is better} \\ \midrule

Finetune-0               & \xmark                           & \xmark                             & $0-80$ m &0.102 & 0.700 & 4.681 & 0.200     & 0.872         & 0.954         & 0.978         \\
Fintuned\_synthesis\_200 & \checkmark                            & \xmark                             & $0-80$ m & 0.100 & 0.682 & 4.515 & 0.195     & 0.879         & 0.957         & 0.979         \\
Fintuned\_synthesis\_700 & \checkmark                            & \xmark                             & $0-80$ m & 0.099 & 0.672 & 4.593 & 0.194     & 0.879         & 0.957         & 0.979         \\ \midrule
Finetuned\_stereo\_gt\_right\_0   & \xmark                           & \xmark & $0-80$ m & 0.062 & 0.424 & 3.677 & 0.164     & 0.939         & 0.968         & 0.981         \\
Finetuned\_stereo\_gt\_right\_200   & \xmark                           & \checkmark                             & $0-80$ m & 0.065 & 0.452 & 3.844 & 0.168     & 0.933         & 0.967         & 0.981         \\
Finetuned\_stereo\_gt\_right\_700   & \xmark                           & \checkmark                             & $0-80$ m & 0.053 & 0.382 & 3.400 & 0.144     & 0.947         & 0.975         & 0.986         \\ \midrule
Finetune-200             & \checkmark                            & \checkmark                             & $0-80$ m & 0.100 & 0.670 & 4.437 & 0.192     & 0.882         & 0.958         & 0.979         \\
Finetune-500             & \checkmark                            & \checkmark                             & $0-80$ m & 0.094 & 0.635 & 4.275 & 0.179     & 0.889         & 0.964         & 0.984         \\
Finetune-700             & \checkmark                            & \checkmark                             & $0-80$ m & 0.094 & 0.626 & 4.252 & 0.177     & 0.891         & 0.965         & 0.984         \\
	\bottomrule
	\end{tabular}
	\end{adjustbox}
	\vspace{-5pt}
	\caption{Quantitative results of different variants of our proposed method on the test set of the KITTI Raw dataset used by Eigen \etal~\cite{eigen2014depth} at the cap of 80m. ``FT VSN'' denotes whether the view synthesis network has been finetuned in an end-to-end fashion, while ``FT SMN'' denotes whether the stereo matching network has been finetuned in an end-to-end fashion. Top three rows: comparisons of different view synthesis network settings. Middle three rows: comparisons of different stereo matching network settings. Bottom three rows: empirical comparisons by different number of training samples. The number in the method names means the number of samples to finetune the network.}
	\label{table:kitti_ablation}
\vspace{-10pt}
\end{table*}

\subsection{Comparisions with state-of-the-art methods}
Next, results on the KITTI Eigen split dataset are reported when right images are predicted by our view synthesis network. Results are compared to six recent baseline methods as showed in Table~\ref{table:kitti_res}, \cite{eigen2014depth,liu2014discrete}  are supervised methods, \cite{kuznietsov2017semi} is a semi-supervised method, and \cite{godard2016unsupervised,zhou2017unsupervised,garg2016unsupervised} are unsupervised methods. Our proposed method is also a semi-supervised method. 

\textbf{Result without end-to-end finetuning:}
After the training of both networks converged, we directly feed the right image synthesized by the view synthesis network to the stereo matching network to predict the depth for the given left images. The result is reported in Table \ref{table:kitti_res}.  

As one can see, even without finetuning the whole network in KITTI dataset, our method performs better than the unsupervised method~\cite{godard2016unsupervised}, and gets comparable performance with the state-of-the-art semi-supervised method~\cite{kuznietsov2017semi}. The performance achieved by our method demonstrates that decoupling the problem of monocular depth estimation into two separate sub-problems is simple yet effective by explicitly enforcing geometrics constraints, which is critical for estimating depth from images.

\textbf{Result with end-to-end finetuning:}
We further finetune the whole system with a small amount of training data from KITTI Eigen split training set, \ie 700 training samples. The left, right images and the depth images are used as training samples to our proposed method.

\begin{figure*}[t!]
\centering
	\includegraphics[width=1.0\linewidth]{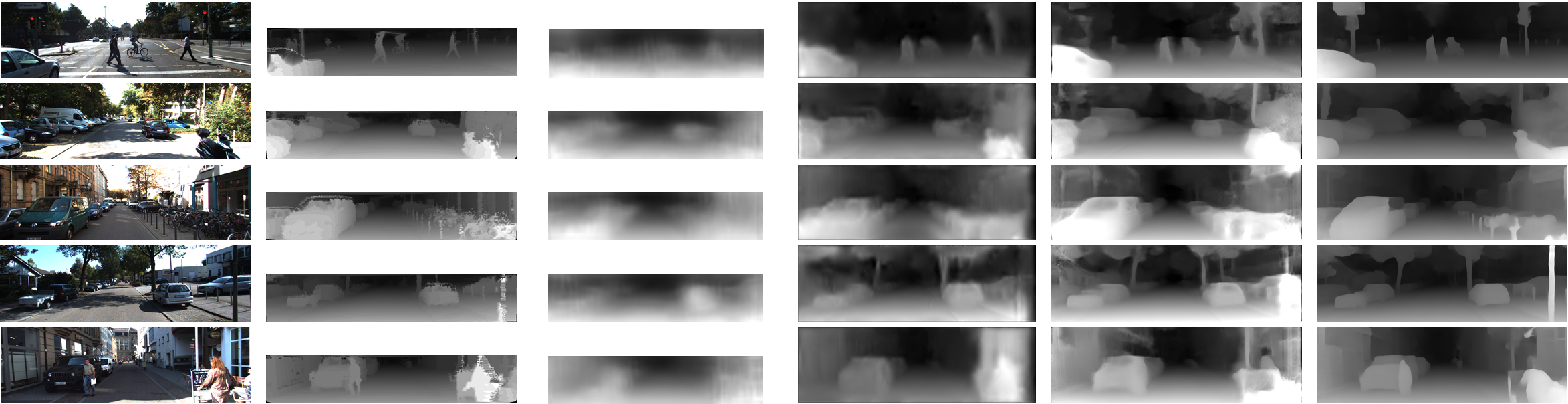}
	\begin{subfigure}[b]{0.16\linewidth}
    \caption{Input}
  	\end{subfigure}
  	\begin{subfigure}[b]{0.16\linewidth}
    \caption{Ground-truth}
  	\end{subfigure}
	\begin{subfigure}[b]{0.16\linewidth}
    \caption{Eigen \etal~\cite{eigen2014depth}}
  	\end{subfigure}
  	\begin{subfigure}[b]{0.16\linewidth}
    \caption{Garg \etal~\cite{garg2016unsupervised}}
  	\end{subfigure}
  	\begin{subfigure}[b]{0.16\linewidth}
    \caption{Godard \etal~\cite{godard2016unsupervised}}
  	\end{subfigure}
  	\begin{subfigure}[b]{0.16\linewidth}
    \caption{Ours}
  	\end{subfigure}
  	\vspace{-12pt}
	\caption{Qualitative results on the KITTI Eigen test set. Sparse ground-truth labels have been interpolated for visualization.
Note that the prediction of our method can better separate the background and foreground or different entities close to each other. Also, our results are crisper and neater. In addition, we are doing better on the objects such as trees, poles, traffic sign and pedestrians, whose depth are generally hard to be inferred accurately.}
	\label{fig:qualitative-reults}
	\vspace{-10pt}
\end{figure*}

The results are reported in Table~\ref{table:kitti_res}, as one can see, our method outperforms all compared methods, with ARD metric reduced by \textbf{17.5\%} compared with Godard \etal~\cite{godard2016unsupervised} and \textbf{16.8\%} compared with Kuznietsov \etal~\cite{kuznietsov2017semi} at the cap of 80 m. Our proposed method performs the best for almost all metrics. It shows that end-to-end training further optimizes the collaboration of these two sub-networks and it leads to the state-of-the-art result. Qualitative comparisons are shown in Figure~\ref{fig:qualitative-reults}. Our proposed method also achieves much more visually accurate estimations than the compared methods.

\subsection{Analyzing the function of two sub-networks after end-to-end training}
In this section, we analyze the function of two sub-networks after end-to-end training. If the end-to-end training breaks the origin functionality of the two sub-networks but the overall performance increases, the whole network would be overfitted to the KITTI dataset, which will make it hard to generalize to other datasets or scenes. To examine the function of two sub-networks, we conduct the following two groups of experiments.

\textbf{Analyzing function of view synthesis sub-network:}
We replaced the stereo matching sub-network in the finetuned network with the one before finetuneing. Since pre-trained stereo matching sub-network is only pre-trained to complete the stereo matching task using real left-right pairs, if after replacing, the whole network could still get good performance in the task of single image depth estimation, the origin functionality of the view synthesis network after the finetuning process could still be retained.

The results are reported in top three rows of Table~\ref{table:kitti_ablation}, denoted as ``Finetuned\_synthesis\_K'', where K represents the number of training samples. As one can see from Table~\ref{table:kitti_ablation}, the results by ``Finetuned\_synthesis\_K'' outperform the method without finetune. From another perspective, the average PSNR between synthesized views and ground truth views in test set increases from 21.29dB to 21.32dB after finetuning. The preservation of functionality may be due to the reason that during the finetuning process, the stereo matching sub-network acts as another loss to better constrain the view synthesis sub-network to generate geometric-reasonable right images. 

\vspace{-4pt}
\begin{table}[!htp]
	\centering
	\footnotesize
	\begin{adjustbox}{max width=1.0\textwidth}
	\renewcommand{\arraystretch}{0}
	\begin{tabular}{@{}l|c|c|c|c|ccc|ccc@{}}
	\toprule
\multicolumn{1}{l|}{Experiment} & \multicolumn{1}{c|}{cap} & ARD   & SRD   & RMSE   & RMSE(log) \\ \midrule
Our\_Best  & $0-80$ m &0.094 & 0.626 & 4.252 & 0.177 \\ 
Kuznietsov \cite{kuznietsov2017semi}  & $0-80$ m &0.113 & 0.741 & 4.621 & 0.189 \\ \midrule
Prob\_Disp  & $0-80$ m & 0.212 & 2.075 & 6.314 & 0.294 \\ \midrule
NoKitti200\_BF  & $0-80$ m &0.119 & 0.969 & 5.079 & 0.207 \\ 
NoKitti200\_AF & $0-80$ m &0.101 & 0.673 & 4.425 & 0.176 \\
	\bottomrule
	\end{tabular}
	\end{adjustbox}
	\vspace{-4pt}
	\caption{Additional experimental results. Upper part is our best result and the previous state-of-the-art result. Middle part shows the result directly calculated from the probabilistic disparity map obtained in our view synthesis network. Lower part shows the results before and after finetuning without 200 high-quality KITTI labels.}
	\label{table:addResults}
\vspace{-6pt}
\end{table}

\textbf{Analyzing function of stereo matching sub-network:}
In order to validate the function of stereo matching sub-network after end-to-end training, we test the stereo matching performance of the finetuned stereo matching sub-network by providing the true left and right image as inputs to predict the depth.

\begin{figure*}[t!]
\centering
	\includegraphics[width=1.0\linewidth]{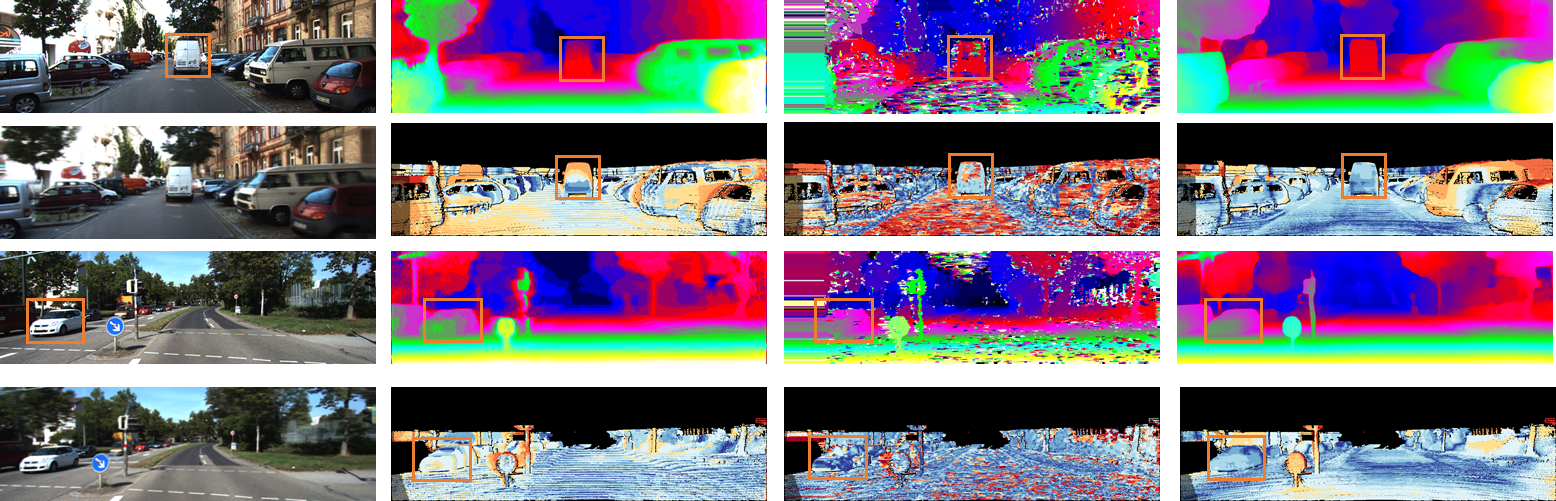}
    \begin{subfigure}[b]{0.24\linewidth}
    \caption{Input image}
  	\end{subfigure}	
	\begin{subfigure}[b]{0.24\linewidth}
    \caption{Godard \etal~\cite{godard2016unsupervised}}
  	\end{subfigure}
  	\begin{subfigure}[b]{0.24\linewidth}
    \caption{OCV-BM}
  	\end{subfigure}
  	\begin{subfigure}[b]{0.24\linewidth}
    \caption{Ours}
  	\end{subfigure}
  	\vspace{-14pt}
	\caption{Empirical study on the qualitative comparisons on KITTI 2015 Stereo test set. The figures from left to right correspond to the input left images, estimated disparity maps or error maps by Godard \etal ~\cite{godard2016unsupervised}, block matching, and our method respectively. And the second and fourth rows are the error maps while the estimated disparity maps are plotted above each error maps, the synthesized right views are also presented in the first column. The error map uses the log-color scale described in~\cite{Menze2015CVPR}, depicting correct estimates in \textcolor{blue}{blue} and wrong estimates in \textcolor{red}{red} color tones. Best view in color.}
	\label{fig:stereo-comparison}
	\vspace{-12pt}
\end{figure*}

The results are provided in the middle three rows of Table~\ref{table:kitti_ablation}, denoted as ``Finetuned\_stereo\_gt\_right\_K''. As shown in Table~\ref{table:kitti_ablation}, ``Finetuned\_stereo\_gt\_right\_200'' performs slightly worse than ``Finetuned\_stereo\_gt\_right\_0'', this may be due to the reason that the finetuning process has forced the stereo matching sub-network to better fit on the imperfect synthesized right images. However, ``Finetuned\_stereo\_gt\_right\_700'' outperforms the pre-trained stereo matching sub-network. The high performance of stereo matching results clearly demonstrates the stereo matching network still maintains its functionality after end-to-end finetuned. 

\begin{figure}[]
\centering
	\includegraphics[width=0.95\linewidth, height=0.5\linewidth]{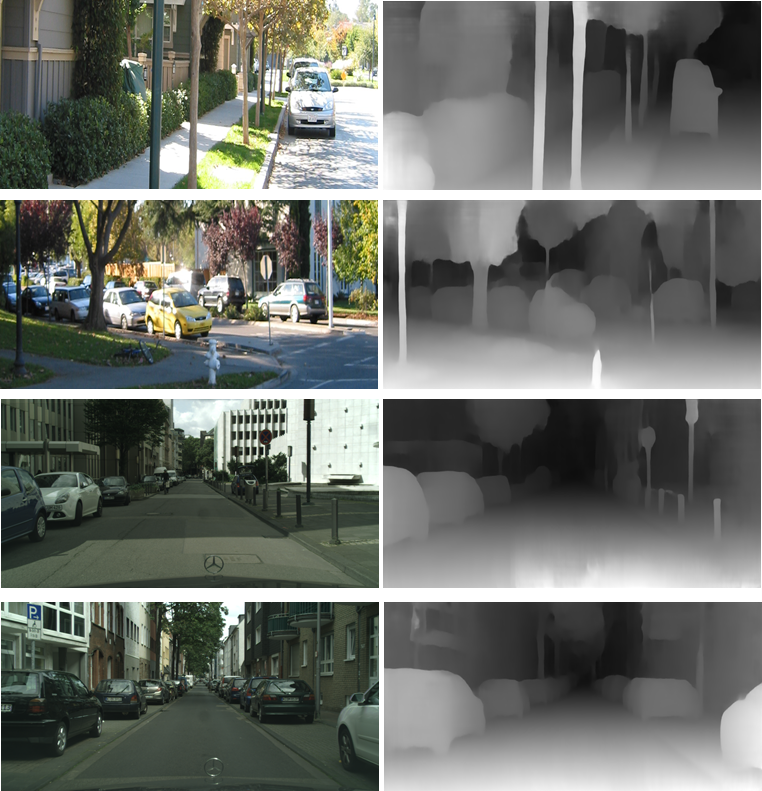}
	\vspace{-5pt}
	\caption{Qualitative results on Make3D dataset \cite{Saxena09make3D} (top two rows) and Cityscapes dataset \cite{Cordts2016Cityscapes} (bottom two rows).}
	\label{fig:generalization}
	\vspace{-15pt}
\end{figure}

Combining the above two experiment groups, we could conclude that after end-to-end training, the two sub-modules collaborate more effectively while preserving their individual functionalities. This may imply that our proposed method could generalized well to other datasets. Some qualitative results on Cityscape dataset~\cite{Cordts2016Cityscapes} and Make3D dataset~\cite{Saxena09make3D} are shown in Figure~\ref{fig:generalization}, which are estimated by our method finetuned in KITTI dataset. The results demonstrate the generalization ability of our proposed method on unseen scenes.

\subsection{Primitive disparity obtained in the view synthesis network} 
Our view synthesis network produces a primitive disparity in order to do the rendering. The middle part in table \ref{table:addResults} shows the estimation accuracy calculated from this probabilistic disparity map. We can see the result is much inferior to the final result of our proposed method. It shows our approach indeed makes a great improvement over the primitive disparity.

\subsection{Analyzing the effect of training sample number} 
To study the effectiveness of our proposed method, we also evaluate our proposed method finetuned by different numbers of samples, \ie, 0, 200, 500, 700, named as ``Finetune-K''. Note that, when K equals to 0, finetuning is not performed on the whole network. 

The results are reported in the bottom three rows of Table~\ref{table:kitti_ablation}. As one can see from the results, more end-to-end finetuning samples could achieve higher performance, and our proposed method could outperform previous state-of-the-art methods by a clear margin only using 700 samples to finetune the whole network. 

\subsection{Use of 200 high-quality KITTI labels} 
As described before, we use 200 high-quality KITTI labels to optionally finetune the stereo matching network. In the lower part of table \ref{table:addResults}, we present the result without these labels before and after finetune(\_BF\&\_AF). We can see that without seeing any real disparity from KITTI, our method already gets promising results. After finetuning without those high-quality labels, our method still beats the current state-of-the-art method. These high-quality labels, in fact, increase the capacity of the model to a certain extent, but without them, our method still makes an improvement under the same condition.

\begin{table}
\centering
\begin{tabular}{|l|c|c|c|}
\hline
Method       & D1-bg & D1-fg & D1-all \\ \hline
Godard \etal~\cite{godard2016unsupervised} & 27.00  & 28.24 & 27.21  \\ \hline
OCV-BM       & \textbf{24.29} & 30.13 & 25.27  \\ \hline
Ours         & 25.18 & \textbf{20.77} & \textbf{24.44} \\ \hline
\end{tabular}
\vspace{-5pt}
\caption{Quantitative results on the test set of the KITTI 2015 Stereo Benchmark~\cite{Menze2015CVPR}. Best results are shown in bold. The number is the percentage of erroneous pixels, and a pixel is considered to be correctly estimated if the disparity is within 3px compared to the ground-truth disparity. Our method has already surpassed the stereo matching method, \ie Block Matching method. }
\label{table:kitti-stereo}
\vspace{-10pt}
\end{table}

\subsection{Comparison with stereo matching method}
In this section, the comparisons with the proposed approach for depth estimation from single images and stereo matching method from stereo images are presented. 
The results are summarized in Table~\ref{table:kitti-stereo}. As one can see, our method is the first single image depth estimation approach that surpasses the traditional stereo matching method, \ie block matching method denoted as ``OCV-BM'' in the table. Exemplar visual results are shown in Fig.~\ref{fig:stereo-comparison}. Because the block matching method directly using low-level image feature to search the matched pixels in the left and right images, the disparity maps predicted by the block matching method are usually noised, which greatly degrades its performance, but the results are still geometrically correct. The geometric reasoning capacity is built in our network and high-level image feature is processed in the deep learning network, these two reasons enable our method to outperform the stereo matching method. Due to the miss of explicit geometric constraints in Godard \etal~\cite{godard2016unsupervised}, its method gets sub-optimal results. Better performance of our method can be seen from the box regions in the figure.

\section{Conclusion}
In this work, we propose a novel perspective to tackle the problem of monocular depth estimation. We show for the first time that this problem can be decomposed into two problems, namely a view synthesis problem and a stereo matching problem. We explicitly encode the geometric transformation within both networks to better tackle the problems individually. Collectively training the whole pipeline results in an overall boost and we prove that both networks are able to preserve their original functionality after end-to-end training. Without using a large amount of expensive ground truth labels, we outperform all previous methods on a monocular depth estimation benchmark. Remarkably, we are the first to outperform the stereo blocking matching algorithm on a stereo matching benchmark using a monocular method.
            
{\small
\bibliographystyle{ieee}
\bibliography{egbib}
}

\end{document}